\pdfoutput=1

\documentclass[11pt]{article}
\usepackage{amsmath}
\usepackage{pgfplots}
\pgfplotsset{compat=1.18} 
\usepackage{amssymb}
\usepackage{verbatim}
\usepackage[final]{acl}
\usepackage{times}
\usepackage{latexsym}

\usepackage{algorithm}
\usepackage{algorithmic}

\definecolor{darkblue}{RGB}{0, 0, 139}

\usepackage[most]{tcolorbox}
\usepackage{listings}
\usepackage{booktabs}
\usepackage{multirow}
\usepackage{colortbl}
\usepackage{xcolor}          
\definecolor{aclblue}{RGB}{235, 240, 250} 
\definecolor{darkblue}{RGB}{60, 80, 120}  

\definecolor{aclblue}{RGB}{230, 240, 255}   
\definecolor{darkblue}{RGB}{30, 60, 120}    

\newtcolorbox{promptbox}[1][]{%
  enhanced,
  breakable,
  colback=gray!2,          
  colframe=gray!20,        
  arc=0mm,
  boxrule=0.4pt,           
  leftrule=2.5pt,          
  coltitle=darkblue,       
  fonttitle=\bfseries\sffamily\large,
  fontupper=\ttfamily\small, 
  title={#1},
  left=12pt, right=12pt, top=10pt, bottom=10pt,
  before skip=1.5ex,
  after skip=1.5ex,
  borderline west={2.5pt}{0pt}{darkblue},
  line width=0.5pt,
  before upper={\linespread{1.1}\selectfont}
}

\usepackage{listings}
\usepackage{algorithm}
\usepackage{algorithmic}
\usepackage{epigraph}

\definecolor{mygrey}{rgb}{0.8, 0.8, 0.8}

\usepackage[T1]{fontenc}

\usepackage[utf8]{inputenc}

\usepackage{microtype}

\usepackage{inconsolata}

\usepackage{graphicx}

\author{
  \textbf{Xin Xie}\textsuperscript{1}\thanks{~~Equal contribution.},
  \textbf{Dongyun Xue}\textsuperscript{2}\footnotemark[1],
  \textbf{Wuguannan Yao}\textsuperscript{1},
  \textbf{Mingxiao Feng}\textsuperscript{2},
  \textbf{Wengang Zhou}\textsuperscript{2}, \\
  \textbf{Xiang Qi}\textsuperscript{1},
  \textbf{Houqiang Li}\textsuperscript{2},
  \textbf{Peng Zhang}\textsuperscript{1}\thanks{~~Corresponding author.} \\
  \textsuperscript{1}Ant Digital Technologies, Ant Group \\
  \textsuperscript{2}Hefei Comprehensive National Science Center, Hefei, China \\
  \texttt{\{xinyuan.xx, yaowuguannan.ywgn, qixiang.qx, minghua.zp\}@antgroup.com}, \\
  \texttt{\{andyxue, fengmx\}@mail.ustc.edu.cn}, \texttt{\{zhwg, lihq\}@ustc.edu.cn}
}

\title{SGA-MCTS: Decoupling Planning from Execution via Training-Free Atomic Experience Retrieval}

\begin{document}
\maketitle

\begin{abstract}
LLM-powered systems require complex multi-step decision-making abilities to solve real-world tasks, yet current planning approaches face a trade-off between the high latency of inference-time search and the limited generalization of supervised fine-tuning.
To address this limitation, we introduce \textbf{SGA-MCTS}, a framework that casts LLM planning as non-parametric retrieval. Offline, we leverage Monte Carlo Tree Search (MCTS) to explore the solution space and distill high-fidelity trajectories into State-Goal-Action (SGA) atoms. These atoms are de-lexicalized primitives that abstract concrete entities into symbolic slots, preserving reusable causal logic while discarding domain-specific noise.
Online, a retrieval-augmented agent employs a hybrid symbolic-semantic mechanism to fetch relevant SGAs and re-ground them into the current context as soft reasoning hints.
Empirical results on complex benchmarks demonstrate that this paradigm enables frozen, open-weights models to match the performance of SOTA systems (e.g., GPT-5) without task-specific fine-tuning.
By effectively amortizing the heavy computational cost of search, SGA-MCTS achieves System 2 reasoning depth at System 1 inference speeds, rendering autonomous planning both scalable and real-time feasible.
\end{abstract}

\begin{figure*}[t] 
    \centering
    \includegraphics[width=0.99\linewidth]{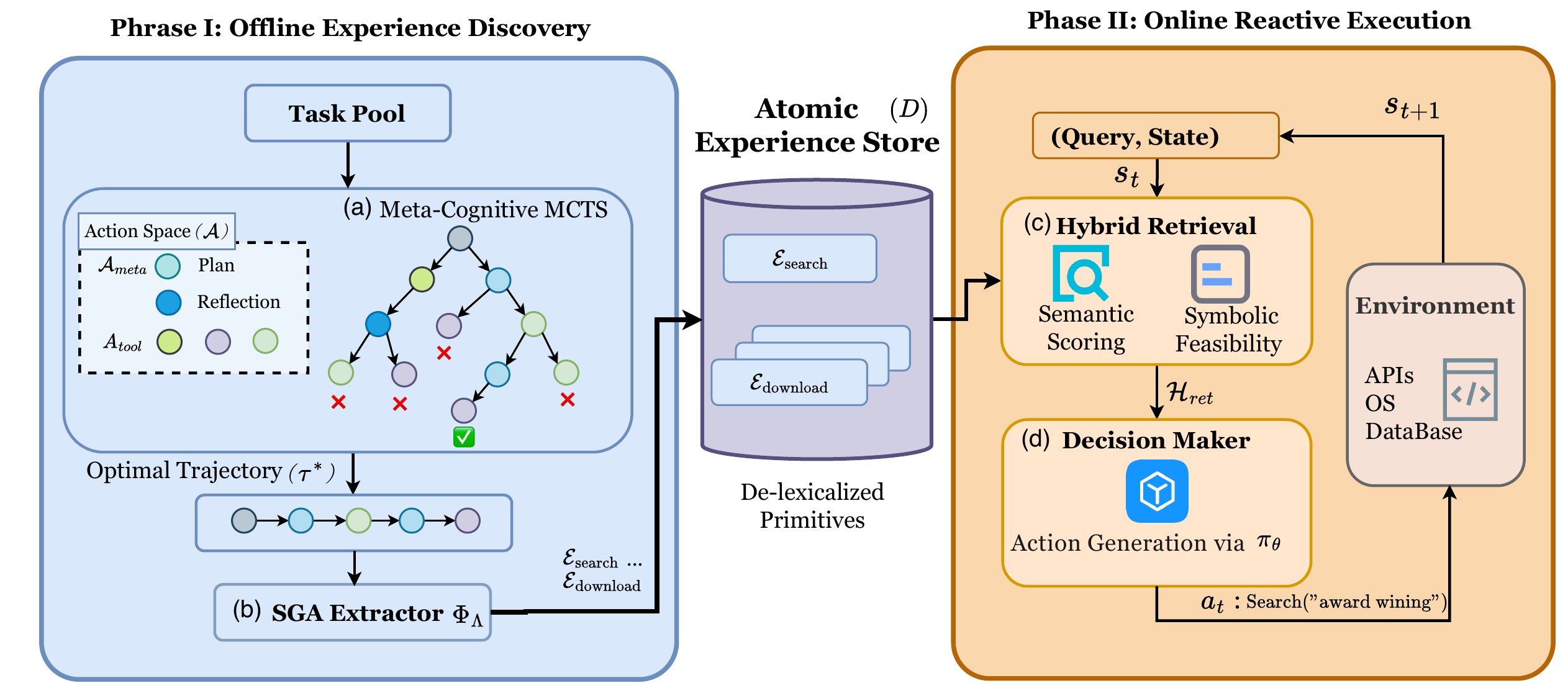} 
    \caption{SGA-MCTS Framework Architecture. (a) MCTS exploration discovers optimal reasoning paths. (b) Valid trajectories are distilled into de-lexicalized SGA atoms. (c) Relevant SGAs are retrieved as soft hints based on the current state. (d) The Decision Maker grounds these hints to generate the final action $a_t$.}
    \label{fig:framework}
\end{figure*}
\section{Introduction}

LLM-based autonomous agents increasingly leverage external tools to solve complex, multi-step problems \cite{schick2023toolformer,qin2023toolllm,wang2024tools,qu2025tool,feng2025retool}, extending capabilities beyond plain text generation.
This transformation has enabled sophisticated capabilities, from booking flights to analyzing scientific data, by grounding language in executable environments.
However, as task complexity grows—demanding long-horizon planning, multi-step dependencies, and dynamic error recovery—agents face an increasingly severe dilemma.
On one hand, they can employ inference-time search methods \cite{yao2023tree,zhou2024language,snell2024scaling} to achieve deep, strategic reasoning, but at the cost of high latency that renders them impractical for interactive applications.
On the other hand, they can embed reasoning patterns into model parameters via supervised fine-tuning, but this suffers from "parametric rigidity": any adaptation to new tool schemas or domain logic demands expensive retraining and risks catastrophic forgetting \cite{masterman2024landscape,chen2025atlas,schick2023toolformer}.
This constraint has created a critical bottleneck, forcing practitioners to choose between depth and deployability.

To address this challenge, we formulate learning as non-parametric experience curation rather than implicit weight adaptation.
Our framework is built on the insight that complex reasoning, despite surface variability, structurally decomposes into recurring, logic-invariant atoms.
For instance, a refund operation adheres to the same abstract protocol regardless of the user ID; similarly, a multi-hop query follows an identical retrieval-verification loop irrespective of the target entities.
While prior memory-based agents attempt to exploit this repetition via monolithic trajectory retrieval \cite{shinn2023reflexion,zhao2024expel,zhang2025agent}, they suffer from contextual rigidity: slight deviations in entity values or schemas often render retrieved plans brittle or irrelevant.
Our approach overcomes this limitation by distilling execution traces into State-Goal-Action (SGA) atoms---symbolic primitives that isolate causal logic from surface details.
By de-lexicalizing concrete entities into typed slots (e.g. \texttt{<ID>}), we transform raw episodes into a composable algebra of reasoning skills, enabling robust transfer across novel tasks and unseen tool ecosystems.

We implement this insight via a two-phase architecture aligned with the dual-process theory of cognition, distinguishing between deliberative planning (System 2) and reactive execution (System 1).
In the offline discovery phase, we employ MCTS as a data generator rather than a direct policy optimizer.
By augmenting the search space with meta-cognitive operators for goal decomposition and error recovery, we extensively explore the state space to identify optimal reasoning trajectories.
These trajectories are subsequently condensed into the SGA store via schema-guided abstraction.
In the online execution phase, the agent operates as a retrieval-augmented generator.
Instead of conducting online search with expensive token cost, it utilizes a hybrid symbolic-semantic retrieval mechanism to fetch relevant SGAs, which are then integrated into the current context as soft reasoning hints.
This framework effectively amortizes the computational cost of planning, enabling the agent to approximate the strategic depth of search-based methods with the inference latency of standard generation.

In contrast to RAG systems that retrieve static facts or memory agents bound by rigid trajectory replay \cite{zhao2024expel,ouyang2025reasoningbank,G-Memory}, SGA-MCTS facilitates the dynamic recombination of abstract reasoning patterns. This flexibility provides two critical advantages: zero-shot generalization to unseen tools via logic re-grounding, and adaptive reasoning depth controlled by context richness.
On complex benchmarks requiring multi-hop dependency resolution, this approach yields substantial gains: a standard open-weights 8B with non-thinking mode achieves a 44.79\% success rate—a 13.86\% absolute improvement over its zero-shot baseline—approaching the performance of proprietary systems like GPT-5 without a single parameter update.
Furthermore, by condensing raw MCTS explorations by $6.9\times$ into reusable atoms, we demonstrate that the heavy computational cost of deep reasoning can be effectively amortized.

Our contributions are:

\begin{enumerate}
    \item \textbf{Amortized Efficiency:} We decouple strategic planning from execution. This eliminates inference-time search overhead and reduces token consumption by $\sim$2,080 tokens per task (76\% reduction) compared to reasoning-heavy baselines, granting the agent System 2 depth at System 1 cost.
    
    \item \textbf{De-lexicalized SGA Abstraction:} We introduce State-Goal-Action atoms that distill raw trajectories into symbolic primitives. This representation achieves a 6.9x compression rate and enables robust zero-shot generalization to unseen toolsets by extracting reusable causal logic from domain specific trajectories.
    
    \item \textbf{Parameter-Efficient Generalization:} We show that intelligent retrieval bridges the gap between model sizes. SGA-MCTS facilitates an 8B model to achieve a +13.86\% gain, effectively outperforming a 32B baseline in a completely training-free manner.
\end{enumerate}

\section{Related Works}

\paragraph{LLM Agents and Tool Use.}
LLMs have evolved from passive generators to proactive agents via external tools \cite{schick2023toolformer, qin2023toolllm,wolflein2025llm,li2025deepagent,li2024agent}. While reasoning-action interleaving (e.g., ReAct \cite{yao2022react}) improves grounding, greedy strategies often fail in long-horizon tasks due to error propagation. Unlike supervised fine-tuning approaches \cite{zhu2025knowagent, masterman2024landscape,liu2024toolace,lin2024hammer} that suffer from "parametric rigidity," SGA-MCTS adopts a non-parametric, training-free paradigm via retrievable experience, enabling zero-shot adaptation without models' parametric updates.

\paragraph{Planning and Inference Search.}
To overcome greedy shortsighted, "test-time scaling" methods \cite{treeofthought,muennighoff2025s1,zhou2024language,erdogan2025plan} introduce deliberate search. However, they incur high latency.
SGA-MCTS address this by shifting computationally intensive search to an offline discovery phase (System 2), allowing the online agent to execute a lightweight, retrieval-augmented policy (System 1) with negligible latency.

\paragraph{Agent Memory.}
Memory mechanisms evolve agents into lifelong learners \cite{zhang2024survey,fang2025memp,yan2025general,zhang2025memevolve,kang2025memory}. While recent frameworks store structured trajectories \cite{G-Memory, ouyang2025reasoningbank}, they often rely on holistic trajectory retrieval, leading to contextual rigidity \cite{zhang2025agent}. SGA-MCTS overcomes this via de-lexicalized atomization, distilling trajectories into abstract $(State, Goal) \rightarrow Action$ primitives to enable compositional generalization.
\section{Methodology}
\label{sec:method}

We propose SGA-MCTS, a framework that decouples planning from execution via training-free atomic experience retrieval. As shown in Figure~\ref{fig:framework}, it operates in two phases: (1) Offline Experience Discovery, where MCTS mines optimal reasoning paths and distills them into de-lexicalized State-Goal-Action atoms; and (2) Online Reactive Execution, where the agent retrieves these atoms as soft hints to solve tasks efficiently. We treat the offline discovery phase as a non-parametric learning process, where atomic SGAs serve as explicit policy proxies.

\subsection{Problem Formulation}
\label{sec:problem_formulation}

We formulate tool-use planning as a Goal-Conditioned Markov Decision Process (MDP), $\mathcal{M} = \langle \mathcal{S}, \mathcal{G}, \mathcal{A}, \mathcal{P}, \mathcal{R} \rangle$. At step $t$, the agent observes state $s_t$, aims for goal $g$, executes action $a_t$, and receives reward $r_t$.

\paragraph{Structured State \& Abstraction.}
Unlike flat token sequences, we define a structured state $s_t = \langle h_t, \mathcal{K}_t, g_t \rangle$, comprising the execution history $h_t$, a symbolic known-info tracker $\mathcal{K}_t$, and the current atomic sub-goal $g_t$. 
We employ a State Abstraction Function $\Phi: \mathcal{S} \rightarrow \hat{\mathcal{S}}$ that de-lexicalizes specific entities in $\mathcal{K}_t$ into typed slots (e.g., \texttt{<ID>}), enabling retrieval across disjoint domains.

\paragraph{Hybrid Action Space.}
The action space $\mathcal{A} = \mathcal{A}_{tool} \cup \mathcal{A}_{meta}$ unifies external API calls ($\mathcal{A}_{tool}$) with internal reasoning operators ($\mathcal{A}_{meta}$). The latter includes \texttt{Plan} for decomposition and \texttt{Reflect} for error handling, allowing the agent to modify its logical trajectory without further interacting with the external environment.

\subsection{Objective and Reward}
We treat the atomic experience store $\mathcal{D}$ as a non-parametric policy support.
Our objective is to optimize $\mathcal{D}$ to maximize the expected return $\mathbb{E}_{\tau \sim \pi_{\mathcal{D}}} [R(\tau)]$.
To balance correctness and efficiency, we design a gated reward function:
\begin{equation}
    R(\tau) = \mathbb{1}_{\text{succ}}(\tau) \cdot \left( (1 - \lambda) + \lambda \cdot \frac{1}{1 + |\tau|} \right)
    \label{eq:reward_function}
\end{equation}
where $\mathbb{1}_{\text{succ}}(\tau) \in \{0, 1\}$ is the binary success indicator, and $|\tau|$ denotes the length of the trajectory.
$\lambda \in [0, 1]$ is a hyperparameter balancing the base reward for correctness and the bonus for efficiency.
This multiplicative formulation ensures that failed trajectories always yield zero reward, strictly biasing the subsequent MCTS exploration toward successful reasoning paths with minimal steps.

\subsection{Phase I: Offline Experience Acquisition}
\label{sec:offline_phase}

To circumvent the reasoning limitations of greedy policies, we employ Monte Carlo Tree Search (MCTS) as a high-fidelity data generator. This phase aims to explore the state space extensively and distill optimal trajectories into a generalized reusable format.

\subsubsection{Meta-Cognitive Search}
\label{sec:mcts_expansion}

We construct a search tree where nodes represent structured states $s$ and edges represent actions. Standard MCTS often struggles with the large branching factor of open-ended tool use. To mitigate this, we augment the action space $\mathcal{A}$ with meta-cognitive operators that function as heuristic pruners during the expansion phase:

\begin{itemize}
    \item Plan ($\mathcal{A}_\text{plan}$): Enforces a topological ordering on sub-goals. By decomposing the global goal $g$ into atomic steps $\{g_1, g_2, \dots\}$, it constrains the search to relevant tool subsets, effectively pruning logically invalid branches.
    \item Reflect ($\mathcal{A}_\text{reflect}$): Triggered upon execution failure. It analyzes error feedback to generate counterfactual pivots, enabling the search to escape local optima that trap standard greedy samplers.
\end{itemize}

\paragraph{State-Goal Deduplicated Exploration.} 
A major advantage of using MCTS as our rollout operator lies in its inherent structural deduplication at the $(State, Goal)$ level. Unlike linear path sampling rollout operators (e.g., ReAct), which redundantly explores the same state across independent trajectories, MCTS aggregates visitation statistics at each decision node, effectively viewing the search space as a directed graph.

MCTS efficiently approximates the optimal policy $(s_t, g_t) \rightarrow a^*_t$.
Guided by the Upper Confidence Bound (UCB) criterion, the search prioritizes high-potential branches while pruning suboptimal sub-trees. 
The accumulated $Q$-values—derived from leaf-node evaluations via the gated reward function (Eq. \ref{eq:reward_function})—serve as a quality assurance filter. 
This mechanism ensures that the search converges specifically on actions that are verifiably correct, rather than merely plausible.
This selectivity is critical for our atomic experience store: we discard raw, noisy trajectories in favor of canonical SGA triplets derived solely from high-confidence paths. Consequently, each stored entry represents a verified transition, ensuring the store remains compact and highly distinctive."

\subsubsection{Atomic SGA Extraction via Schema-Guided Abstraction}
\label{sec:sga_extraction}

Upon the convergence of MCTS, the search tree provides a set of high-reward trajectories $\tau^*$. To transform these monolithic sequences into reusable knowledge, we decompose each successful transition into a State-Goal-Action (SGA) triplet. Storing raw transitions (e.g., searching specifically for "award-winning" films) leads to "lexical overfitting." To bridge this gap, we introduce a Schema-Guided Abstraction Function $\Phi_\Lambda$ that distills raw execution data into generalized reasoning "atoms."

For each optimal transition $(s_t, g_t, a^*_t)$ identified by MCTS, we generate an atomic experience $\mathcal{E}_i$:
\begin{equation}
    \mathcal{E}_i = \Phi_\Lambda(s_t, g_t, a^*_t) = \langle \hat{S}, \hat{G}, \hat{A} \rangle
\end{equation}

The abstraction process is governed by three core components:

\begin{enumerate}
    \item State Abstraction ($\hat{S}$): Instead of preserving the full history $h_t$, $\hat{S}$ encapsulates a semantic summary of the context and a symbolic schema $\hat{S}_{sym}$. The latter identifies the entity types currently verified in $\mathcal{K}_t$ (e.g., \texttt{<MOVIE\_QUERY>} or \texttt{<ID>}), which serves as a prerequisite for the action.
    \item Goal Abstraction ($\hat{G}$): The sub-goal $g_t$ (e.g., "Retrieve streaming link") is preserved as the functional intent. This allows the agent to retrieve the atom based on its "intended utility" rather than surface text matching.
    \item Action De-lexicalization ($\hat{A}$): We employ a LLM to identify a selective mask to parameters: arguments matching the domain schema (e.g., \texttt{query}) are replaced by typed slots (\texttt{<QUERY>}), while control literals essential for API behavior  are preserved. This hybrid structure ensures the agent generalizes to novel data while retaining the execution logic discovered via MCTS.
\end{enumerate}


\newcommand{\std}[1]{\scriptsize{$\pm$#1}}
\newcommand{\cc}[1]{\cellcolor{aclblue}#1}

\begin{table*}[t!]
\centering
\small

\renewcommand{\arraystretch}{1.25} 

\setlength{\tabcolsep}{15pt}

\begin{tabular}{ll c c c c}
\toprule
\multirow{2}{*}{\textbf{Backbone}} & \multirow{2}{*}{\textbf{Method}} & \textbf{StableTB} & \textbf{ToolHop} & \textbf{BFCL v3} & \multirow{2}{*}{\textbf{Avg.}} \\
& & \textit{(Complex)} & \textit{(Dependency)} & \textit{(Multi-turn)} & \\

\midrule

\textit{Proprietary} & GPT-5 & 70.20 & 43.52 & 51.68 & 55.13 \\

\midrule

\multirow{3}{*}{\textbf{Qwen3-8B}} 
& ReAct & 11.50 \std{0.80} & 36.94 \std{3.35} & 44.35 \std{3.65} & 30.93 \\
& LangMem & 19.30 \std{2.60} & 39.72 \std{0.56} & 46.07 \std{1.19} & 35.03 \\
& \cc{\textbf{SGA (Ours)}} & \cc{\textbf{43.80} \std{2.10}} & \cc{\textbf{41.87} \std{0.71}} & \cc{\textbf{48.70} \std{1.40}} & \cc{\textbf{44.79}} \\

\addlinespace[0.6em]

\multirow{3}{*}{\textbf{Qwen3-14B}} 
& ReAct & 21.70 \std{2.90} & 36.38 \std{0.16} & 47.47 \std{1.86} & 35.18 \\
& LangMem & 28.50 \std{2.80} & 36.21 \std{0.67} & 48.22 \std{1.13} & 37.64 \\
& \cc{\textbf{SGA (Ours)}} & \cc{\textbf{40.20} \std{4.70}} & \cc{\textbf{46.63} \std{1.21}} & \cc{\textbf{52.33} \std{2.21}} & \cc{\textbf{46.39}} \\

\addlinespace[0.6em]

\multirow{3}{*}{\textbf{Qwen3-32B}} 
& ReAct & 18.20 \std{1.80} & 48.70 \std{0.51} & 53.53 \std{1.85} & 40.14 \\
& LangMem & 24.70 \std{2.50} & 48.24 \std{0.67} & 52.27 \std{1.13} & 41.73 \\
& \cc{\textbf{SGA (Ours)}} & \cc{\textbf{48.20} \std{2.00}} & \cc{\textbf{50.87} \std{0.37}} & \cc{\textbf{54.20} \std{1.20}} & \cc{\textbf{51.09}} \\

\bottomrule
\end{tabular}

\caption{Main Results: Success Rate (\%) comparison across different backbones. We compare SGA-MCTS (highlighted in \colorbox{aclblue}{blue}) against baselines. The table is formatted to span a wider area for better readability.}
\label{tab:main_results}
\end{table*}
\subsection{Phase II: Online Reactive Execution}
\label{sec:online_phase}

In the online phase, the agent transitions into a reactive executor (System 1), operating under strict latency constraints. We forego computationally expensive search in favor of a lightweight retrieve-inject-generate pipeline. We treat the retrieved experiences not as rigid commands or templates to be filled, but as soft reasoning hints that provide a logical prior for the agent's generative process.

\subsubsection{Hybrid Symbolic-Semantic Retrieval}
\label{sec:hybrid_retrieval}

Standard dense retrieval often fails in tool-use scenarios because it prioritizes surface-level semantic similarity while ignoring the hard constraints of execution logic. To address this, we propose a dual-factor scoring mechanism that evaluates candidate experiences across two orthogonal dimensions: semantic relevance and symbolic feasibility.

At each timestep $t$, the agent constructs a query vector $\mathbf{q}_t$ and identifies the set of currently available symbolic slots $\Lambda_{t} = \text{Keys}(\mathcal{K}_t)$ updated via an LLM-based state tracker. The unified relevance score for a candidate experience $\mathcal{E}_i$ in the store $\mathcal{D}$ is:

\begin{equation}
\label{eq:hybrid_score}
\resizebox{0.9\linewidth}{!}{$ 
    \text{Score}(\mathcal{E}_i \mid \mathbf{q}_t, \Lambda_t) = (1 - \beta) \cdot \underbrace{\cos(\mathbf{q}_t, \mathbf{e}_i)}_{\text{Semantic Relevance}} + \beta \cdot \underbrace{\frac{|\Lambda_{t} \cap \hat{S}_{sym}^i|}{|\hat{S}_{sym}^i| + \epsilon}}_{\text{Symbolic Feasibility}}
$}
\end{equation}
where $\beta \in [0, 1]$ modulates the balance between intent alignment and execution grounding. Symbolic feasibility acts as a logical gate, penalizing atoms whose prerequisite parameters are missing from the current state, thereby filtering out unexecutable "hallucinated" plans.

\subsubsection{Action Generation via the Decision Maker}
\label{sec:generative_arbitration}

The Decision Maker functions as a generative synthesizer that grounds abstract reasoning patterns into executable actions. 
Instead of enforcing rigid templates, we inject the top-$k$ retrieved SGA triplets into the agent's context as soft logical priors. 
Leveraging the model's in-context learning capabilities, the Decision Maker performs implicit instantiation: it autonomously maps the symbolic slots in the de-lexicalized atoms (e.g., \texttt{<QUERY>}) to concrete entities derived from the execution history $h_t$.
This allows for adaptation where retrieved logic guides, rather than constrains, the generation process based on real-time observation $o_t$.

The decision process is formally modeled as a conditional generation task:
\begin{equation}
\label{eq:arbitration_formula}
    a_t \sim \pi_\theta(a_t \mid h_t, \mathcal{E}_{\text{ret}})
\end{equation}

This ``Reasoning-as-Retrieval'' paradigm enables the agent to exhibit the strategic depth of offline search at the latency of greedy generation, effectively bypassing the fragility of manual slot-filling or rule-based execution.
\section{Experiments}
\label{sec:experiments}

To validate the effectiveness and efficiency of SGA-MCTS, we conducted extensive evaluations on three diverse benchmarks covering complex tool chaining, embodied decision-making, and multi-turn state tracking.

\subsection{Experimental Setup}

\paragraph{Datasets.} To evaluate our framework's capability to generalize from offline exploration to online execution, we utilize three datasets with specific split strategies for experience store construction (Offline) and evaluation (Online):

\begin{itemize}
    \item StableToolbench \cite{guo2024stabletoolbench}: We adopt a cross-difficulty transfer setting. We utilize the G2 Instruction subset (intermediate complexity) for offline experience discovery and evaluate on the G3 Instruction subset (complex/hard). This tests the agent's ability to extrapolate logic from simpler scenarios to long-horizon tasks.
    
    \item ToolHop\cite{toolhop}: A query-driven benchmark for multi-hop tool use containing 995 complex queries. We randomly sample 50\% of the data for offline exploration to build the SGA store. The remaining 50\% are reserved for evaluation, ensuring the agent must handle unseen query-tool dependencies.
    
    \item BFCL v3 (Multi-turn Base) \cite{bfcl}: A challenging subset of the Berkeley Function Calling Leaderboard focusing on multi-turn dialogue state tracking. We sample only 25\% of the episodes for offline atom extraction and evaluate on the remaining 75\%. This low-resource setting stress-tests the efficiency of our symbolic state tracking mechanism.
\end{itemize}

\paragraph{Baselines.} We compare SGA-MCTS against the following representative paradigms:
\begin{itemize}
    \item ReAct (Zero-shot) \cite{yao2022react}: The most widely used prompting-based baseline. It relies solely on the model's internal parametric knowledge to interleave reasoning and tool calls. This baseline serves to demonstrate the "baseline" capability of the Qwen3 backbone without any external experience guidance.
    \item LangMem \cite{langmem2025}: A representative long-term memory baseline. It extracts and stores key information from interactions to enable future retrieval. Specifically, we adopt its episodic memory implementation, which allows the agent to continuously improve by learning from past experiences. This baseline serves to evaluate the benefits of retrieval-based memory guidance.
\end{itemize}

Consistent with the training-free nature of our approach, we evaluate the Qwen3 family (8B, 14B, and 32B) \cite{yang2025qwen3} in standard non-thinking mode to isolate gains strictly from our retrieval mechanism.
Conversely, we employ GPT-5 in thinking mode as a high-ceiling reference to measure the extent to which pure retrieval can bridge the gap to proprietary reasoning models.

\paragraph{Superiority in Complex Planning.}
SGA-MCTS achieves consistent and substantial improvements across all datasets. On average, our method boosts the performance of Qwen3-8B by 13.86\% absolute (from 30.93\% to 44.79\%). This advantage is particularly pronounced on StableToolBench, the most challenging benchmark involving complex instruction following, where SGA-MCTS achieves a relative improvement of nearly 400\% over the ReAct baseline (43.80\% vs. 11.50\%).

\paragraph{Structured Abstraction vs. Holistic Memory.}
While LangMem improves over ReAct by retrieving past trajectories (35.03\% avg), it still lags significantly behind SGA-MCTS (44.79\% avg). 
The performance gap is widest on StableToolBench (19.30\% for LangMem vs. 43.80\% for SGA). 
This disparity supports our hypothesis regarding "contextual rigidity": LangMem's retrieval of raw trajectories often fails when specific entity values or constraints shift in new tasks. In contrast, SGA's de-lexicalized abstraction isolates reusable causal logic from domain noise, enabling robust generalization even when the surface form of the task changes drastically.

\paragraph{Parameter-Efficient Planning.}
We observe that external memory can serve as a potent alternative to purely parametric scaling. Notably, the Qwen3-8B agent with SGA achieves performance comparable to, and in some cases exceeding, the much larger Qwen3-32B baseline. This suggests that for logic-intensive tasks, retrieving curated reasoning patterns offers a resource-efficient path to high performance, reducing the dependency on massive model size.

\paragraph{Closing the Gap with Proprietary SOTA.}
SGA-MCTS allows open-weights models to approximate closed-source models.
The Qwen3-32B + SGA agent achieves an average success rate of 51.09\%, significantly narrowing the gap with GPT-5 (55.13\%). On the BFCL v3 benchmark, our method actually outperforms GPT-5 (54.20\% vs. 51.68\%), demonstrating that specialized, retrieval-augmented planning can exceed the capabilities of general-purpose frontier models in structured tool-use scenarios.
\subsection{Efficiency and Resilience}
\label{sec:efficiency}

Beyond success rates, Table \ref{tab:hops_merged} evaluates the computational cost and stability on the hardest tasks.

\paragraph{Amortized Cost and Inference Efficiency.} 
The efficiency gains of SGA-MCTS extend beyond simple token savings to a fundamental shift in computational allocation. As detailed in Table \ref{tab:hops_merged}, our method reduces token consumption by 76\% ($\sim$2,080 tokens per task) compared to the \textit{ReAct-Thinking} baseline. 
Traditional "inference-time scaling" approaches (e.g., CoT or online search) incur a linear "reasoning tax" for every query, requiring extensive token generation to traverse the solution space. 
In contrast, SGA-MCTS amortizes this cost into the offline phase. By converting complex reasoning paths into retrievable static assets, the online agent effectively "memorizes" the strategic depth of MCTS. This allows it to achieve System 2-level decision quality with the latency profile of a shallow, greedy executor, making high-intelligence planning viable for latency-sensitive deployment.

\paragraph{Resilience to Reasoning Depth and Drift.} 
Long-horizon planning is notoriously brittle due to the cascading error propagation inherent in autoregressive generation. Table \ref{tab:hops_merged} quantifies this degradation: baseline performance collapses precipitously as task complexity increases, dropping to a mere 15.38\% on chains exceeding 4 hops. 
SGA-MCTS, however, exhibits remarkable resilience, maintaining a robust success rate of 61.54\% in these deep-dependency scenarios. 
This stability stems from the function of retrieved SGA atoms as logic checkpoints.
Instead of relying solely on a volatile context window that drifts over time, the agent re-grounds its logic at every step using validated, de-lexicalized schemas.
This mechanism effectively resets the "reasoning uncertainty" at each hop, preventing the hallucination drift that typically derails long-chain execution.

\begin{table}[h]
\centering
\resizebox{\columnwidth}{!}{
\begin{tabular}{l c c c}
\toprule
\multirow{2}{*}{\textbf{Method}} & \multicolumn{2}{c}{\textbf{Success Rate (\%)}} & \multirow{2}{*}{\textbf{Avg. Tokens}} \\
\cmidrule(lr){2-3} 
 & \textbf{Easy} & \textbf{Hard} &  \\
\midrule
ReAct (Baseline) & 16.67 & 7.69 & 260.54 \\
ReAct - Thinking & 31.25 & 15.38 & 2712.75 \\
\midrule
\rowcolor{aclblue} \textbf{SGA-MCTS (Ours)} & \textbf{43.75} & \textbf{61.54} & \textbf{630.28} \\
\quad \textit{Imp. vs. React-Thinking} & \textcolor{blue}{+12.50} & \textcolor{blue}{\textbf{+46.16}} & \textcolor{teal}{\textbf{-2082.47}} \\
\bottomrule
\end{tabular}
}
\caption{Efficiency on StableToolBench G3. SGA-MCTS achieves significantly higher success rates on hard tasks while maintaining efficient token usage compared to reasoning-heavy baselines.}
\label{tab:hops_merged}
\end{table}

\section{Ablation and Diagnostic Study}
To isolate the sources of improvement, we conduct a fine-grained analysis of the framework's constituents: the topology of offline discovery, the impact of de-lexicalized atoms, and the sensitivity of hybrid retrieval.

\subsection{Quality of Offline Discovery}
The effectiveness of our framework hinges on the MCTS to mine high-quality logic offline.

\begin{table}[h]
\centering
\small
\renewcommand{\arraystretch}{1.05} 
\setlength{\tabcolsep}{3.5pt} 
\resizebox{\linewidth}{!}{
    \begin{tabular}{l c c c c}
    \toprule
    \multirow{2}{*}{\textbf{Backbone}} & \multicolumn{2}{c}{\textbf{Search Topology Metrics}} & \multirow{2}{*}{\textbf{Avg Nodes}} \\
    \cmidrule(lr){2-3}
     & \textbf{Branch. Factor} & \textbf{Avg Depth} & \\
    \midrule
    Qwen3-8B  & 1.25 \scriptsize{$\pm$ 0.24} & 11.98 \scriptsize{$\pm$ 3.81} & 28.9 \\
    Qwen3-14B  & 1.29 \scriptsize{$\pm$ 0.30} & 13.50 \scriptsize{$\pm$ 5.16} & 38.4 \\
    Qwen3-32B & 1.30 \scriptsize{$\pm$ 0.30} & 13.96 \scriptsize{$\pm$ 4.95} & 40.4 \\
    \bottomrule
    \end{tabular}
}
\caption{MCTS Exploration Statistics (BFCL v3). The low \textit{Branch. Factor} ($\approx 1.3$) and high \textit{Depth} ($>11$) indicate efficient, deep reasoning.}
\label{tab:mcts_metrics}
\end{table}
\paragraph{Deep-but-Narrow Exploration.}
Table \ref{tab:mcts_metrics} reveals a "deep-but-narrow" search topology (branching $\approx 1.3$, depth $>11$). This corroborates that our meta-cognitive operators successfully guide exploration through narrow logical corridors, avoiding shallow heuristics. By encapsulating these expensive trajectories into retrievable atoms, we effectively amortize search costs, strictly decoupling reasoning depth from online latency.

\begin{table}[h]
\centering
\resizebox{\linewidth}{!}{
    \renewcommand{\arraystretch}{1.2}
    \begin{tabular}{l c c}
    \toprule
    \textbf{Dataset} & \textbf{ Explored Actions} & \textbf{SGAs ($|\mathcal{D}|$)} \\
    \midrule
    StableToolBench & 2388 & 213 \\
    ToolHop & 10685 & 1560 \\
    BFCL v3 & 1290 & 226 \\
    \bottomrule
    \end{tabular}
}
\caption{Statistics of the atomic experience store. We compare the volume of raw actions explored via MCTS during the offline phase against the final size of the deduplicated, de-lexicalized atomic experience store ($\mathcal{D}$). The reduction highlights the high reusability of the distilled reasoning atoms.}
\label{tab:sga_stats}
\end{table}

\paragraph{High-Density Compression.}
Our targeted exploration strategy yields significant data compaction. As shown in Table \ref{tab:sga_stats}, we consolidate 10,685 raw actions from the ToolHop dataset into just 1,560 reusable atoms—a compression factor of $\sim 6.9\times$. This finding validates that diverse tasks share a common, low-dimensional basis of recurring causal logic. Instead of storing redundant execution traces, SGA-MCTS captures these essential patterns, thereby mitigating the combinatorial explosion of the state space.

\subsection{Retrieval Sensitivity and Hybrid Scoring}

\begin{figure}[h]
    \centering
    \includegraphics[width=0.95\linewidth]{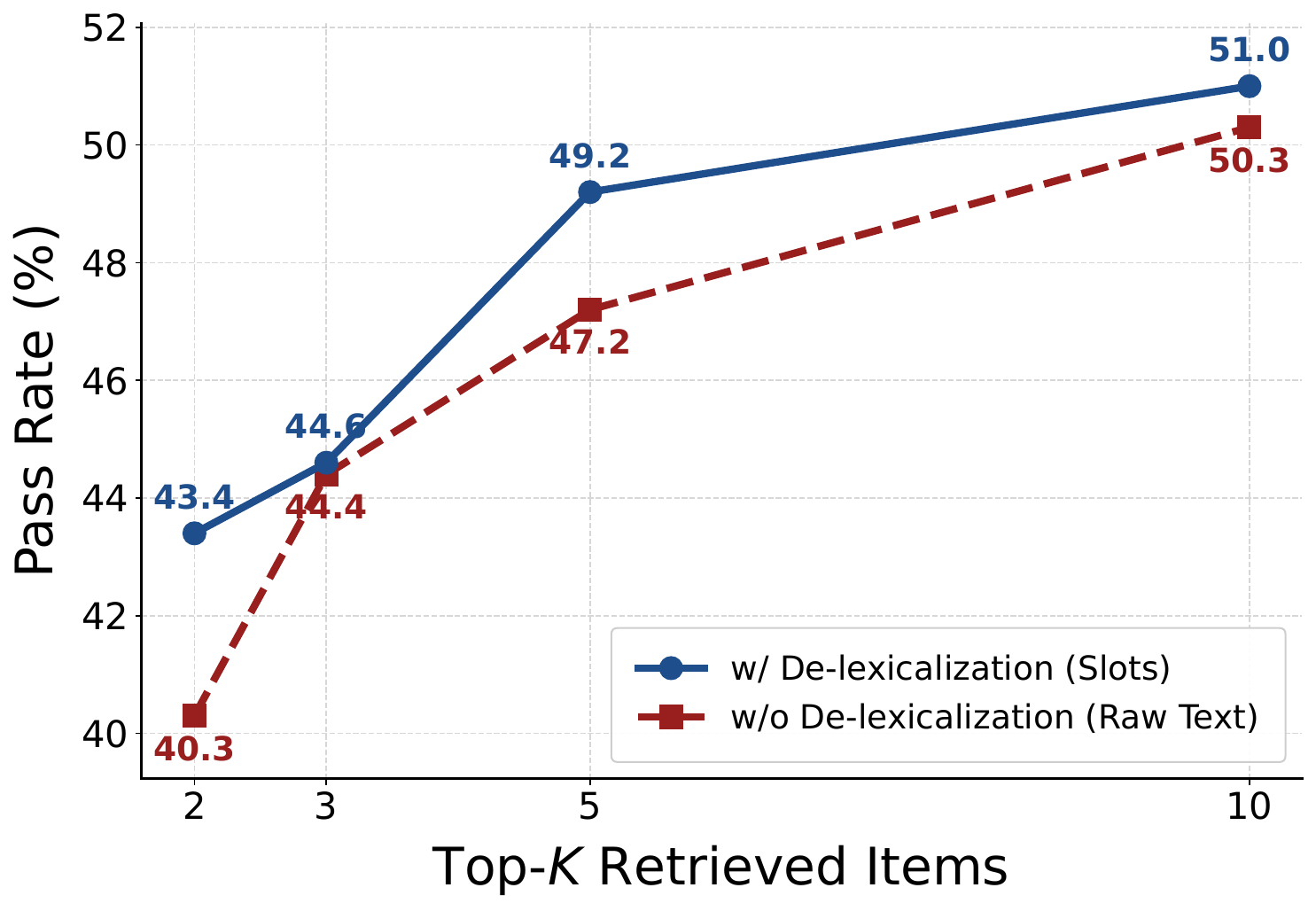}
    \caption{Impact of Retrieval Size ($K$) on StableToolBench. Unlike Raw Text (Red) which degrades due to noise, our De-lexicalized approach (Blue) maintains robust performance as $K$ increases.}
    \label{fig:top_k}
\end{figure}

\paragraph{Top-K Sensitivity.} 
Figure \ref{fig:top_k} shows a clear positive trend: performance consistently improves as $K$ increases. The success rate climbs steadily with more retrieved SGA atoms, indicating that the model effectively uses the richer context to ground its reasoning. There is no performance degradation from $Top-k$ grows, suggesting that more reference examples provide stronger logical guidance.

\paragraph{Necessity of Symbolic Constraints.}
Ablating the symbolic term ($\beta=0$) on Qwen3-8B yields a 1.5\% drop, driven by \textit{precondition hallucinations}--invoking tools without requisite arguments.
As shown in Figure \ref{fig:top_k}, this confirms symbolic feasibility acts as a critical validity gate. 

\subsection{Generalization to Unseen Tools}

We investigate whether SGA achieves robust generalization or merely relies on memorization by correlating performance gains with tool familiarity.

\paragraph{Metric: Tool Familiarity Score.}
To quantify the semantic distribution shift between the offline discovery toolset ($\mathcal{T}_{src}$) and the online evaluation toolset ($\mathcal{T}_{tgt}$), we introduce the Tool Familiarity Score ($\mathcal{S}_{\text{fam}}$).
Unlike rigid overlap statistics, $\mathcal{S}_{\text{fam}}$ operates in the dense embedding space to measure the \textit{semantic proximity} of the testing environment to the source domain.
For each tool $t$ in the target set, we identify its nearest neighbor in the source set and compute the average peak similarity:
\begin{equation}
    \mathcal{S}_{\text{fam}} = \frac{1}{|\mathcal{T}_{tgt}|} \sum_{t \in \mathcal{T}_{tgt}} \max_{t' \in \mathcal{T}_{src}} \cos(\mathbf{e}_t, \mathbf{e}_{t'})
\end{equation}
where $\mathbf{e}_t$ denotes the dense embedding of the tool's functional description.
Intuitively, $\mathcal{S}_{\text{fam}}$ serves as a continuous proxy for \textit{domain novelty}: a score approaching $1.0$ implies an in-distribution setting where the agent can rely on memory, whereas a lower score indicates a high-entropy OOD scenario, demanding the transfer of abstract reasoning logic to semantically distinct tools.

\begin{figure}[h]
    \centering
    \includegraphics[width=1\linewidth]{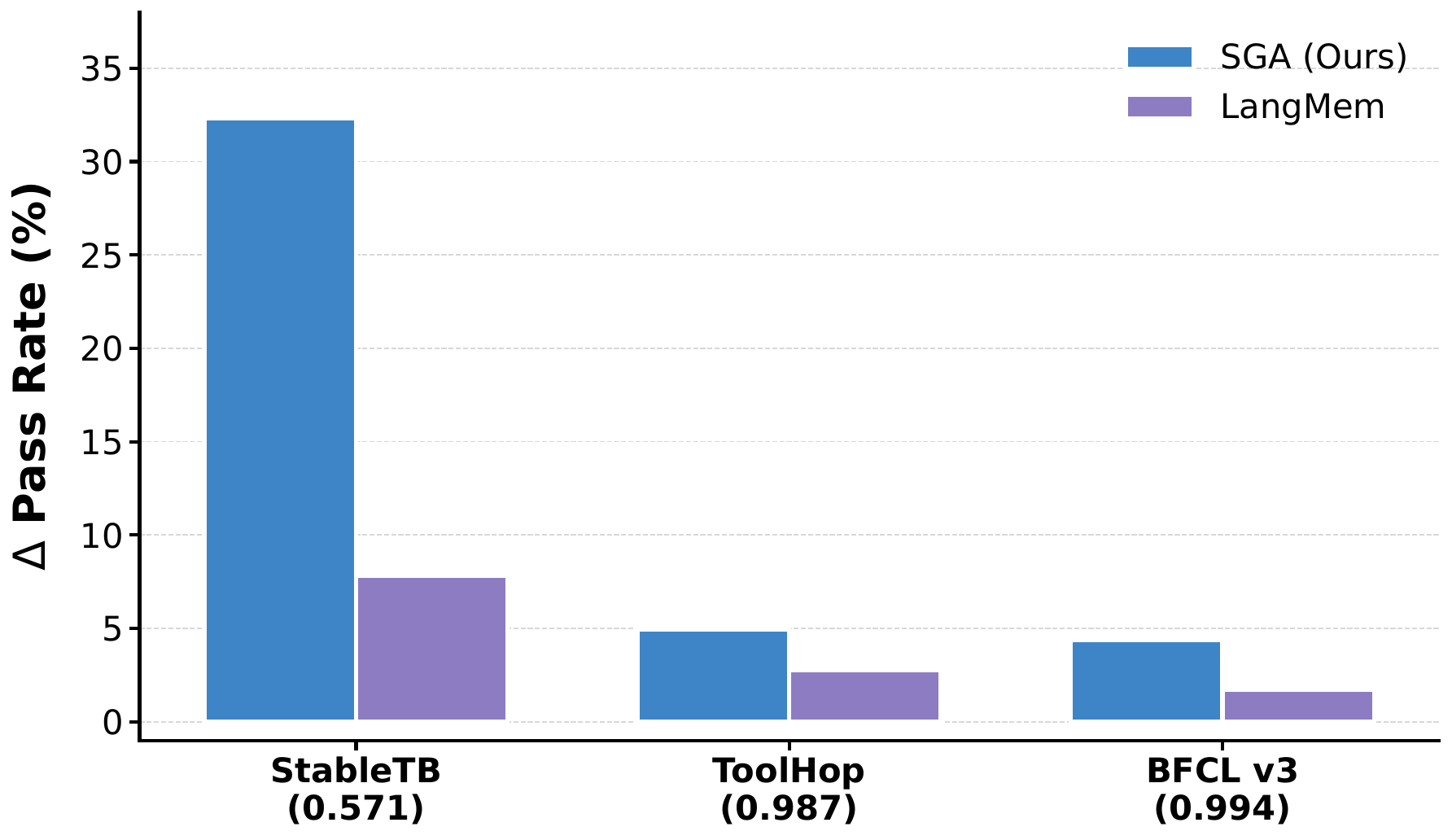}
    \caption{$\Delta$ Pass Rate vs. Dataset (Tool Familiarity). The inverse relationship demonstrates that SGA is most effective in OOD settings (low tool familiarity), showing its strong ability to generalize abstract reasoning logic to unseen tools.}
    \label{fig:delta_familiarity}
\end{figure}

\paragraph{Results Analysis.}
Figure \ref{fig:delta_familiarity} illustrates the performance disparity between SGA-MCTS and LangMem\cite{langmem2025} across varying tool familiarity.
On high-familiarity benchmarks (e.g., BFCL v3, $\mathcal{S}_{\text{fam}} \approx 0.99$), the gap is narrow, as LangMem's retrieval of raw, lexically-matched experience remains effective for seen tools.
However, this gap widens significantly on OOD domains. On StableToolBench ($\mathcal{S}_{\text{fam}} \approx 0.57$), where LangMem expose the limitations due to contextual rigidity, SGA achieves a dominant lead (43.80\% vs. 19.30\% for 8B).
This confirms that while raw memory suffices for reproduction, SGA's de-lexicalized abstraction is essential for OOD generalization, enabling the transfer of reasoning logic to different tool ecosystems.

\section{Conclusion}
\label{sec:conclusion}
We presented SGA-MCTS, a framework that decouples deliberative planning from reactive execution. By amortizing the heavy computational cost of search into an offline phase, we address the limitations of parametric rigidity, recasting planning as the efficient retrieval of de-lexicalized atomic experiences. Our results demonstrate that this non-parametric approach allows frozen, small-scale models to approximate the reasoning depth of proprietary frontier systems without task-specific fine-tuning.
By successfully embedding System 2 reasoning patterns into a retrievable System 1 format, SGA-MCTS offers a scalable path toward interpretable autonomy. We envision this 'Reasoning-as-Retrieval' paradigm as a promising direction for future research, particularly in enabling robust generalization across  dynamic environments.

\section{Limitations}
\label{sec:limitations}

Despite the efficiency gains, SGA-MCTS faces two primary constraints. First, the quality upper bound: the online agent's performance is strictly limited by the fidelity of the offline MCTS exploration. Inaccurate verification during the discovery phase leads to the storage of sub-optimal logic ("low-quality trajectories"), and extremely high-entropy domains may challenge the coverage of our finite experience store.

Second, initialization via seed questions. The construction of the atomic experience store is currently driven by a set of cold-start queries. While our approach efficiently mines optimal reasoning paths (\textit{depth}) within these tasks, the categorical \textit{breadth} of the store is naturally influenced by the diversity of the initial input distribution. Exploring mechanisms for autonomous task proposal (e.g., Active Learning) to expand coverage beyond the seed set remains a promising direction for future research.


\bibliography{custom}

\clearpage
\appendix

\section{Implementation Details}
\label{app:implementation}

\subsection{Model Configuration and Hyperparameters}
To facilitate reproducibility, we detail the specific configurations used in the SGA-MCTS framework. We employ the Qwen3 model family (8B, 14B, and 32B) \cite{yang2025qwen3} as the backbone for all agentic components, encompassing both the offline Planner and the online Decision Maker. 

To balance the trade-off between generative creativity and instruction-following adherence, we enforce a unified decoding strategy across all tasks. Specifically, we set the temperature to $0.6$, with nucleus sampling parameters \texttt{top\_p} $= 0.95$ and \texttt{top\_k} $= 20$. Additionally, we utilize \texttt{min\_p} $= 0$, consistent with the default configuration specified in the model's \texttt{generation\_config.json}. 

These sampling parameters are maintained consistently during the offline MCTS phase to ensure that the distilled experiences remain representative of the model's natural probability distribution. The specific hyperparameters governing the MCTS exploration method and the hybrid retrieval mechanism are systematically summarized in Table \ref{tab:hyperparams}.

\begin{table}[h]
\centering
\small
\renewcommand{\arraystretch}{1.1}

\begin{tabular}{llc}
\toprule
\textbf{Module} & \textbf{Parameter} & \textbf{Value} \\
\midrule
\multirow{4}{*}{Offline MCTS} & Exploration Constant ($c$) & 1.41 \\
 & Max Iterations ($N$) & 50 \\
 & Max Depth & 10 \\
 & Lambda ($\lambda$) & 0.1 \\
\midrule
\multirow{4}{*}{Generation} & Temperature ($T$) & 0.6 \\
 & Top-$P$ / Top-$K$ & 0.95 / 20 \\
 & Min-$P$ & 0.0 \\
 & Max Context Window & 32k \\
\midrule
\multirow{5}{*}{Retrieval} & Semantic Weight ($\alpha$) & 0.7 \\
 & Symbolic Weight ($\beta$) & 0.3 \\
 & Embedding Model & \texttt{bge-m3} \\
 & Smoothing Term ($\epsilon$) & 1e-5 \\
 & Retrieved SGA Top-$k$  & 3 \\
\bottomrule
\end{tabular}
\caption{Hyperparameters for the SGA-MCTS Framework.}
\label{tab:hyperparams}
\end{table}

\subsection{Computational Infrastructure}
All experiments, including the computationally intensive offline MCTS trajectory distillation and the online evaluation benchmarks, were conducted on a high-performance computing cluster. The infrastructure consists of 8 $\times$ NVIDIA A100 (80GB) GPUs. This substantial VRAM capacity is essential for the efficient parallel inference of the Qwen3-32B model. 

For the atomic experience store, we leverage the FAISS library (CPU-optimized build) \cite{douze2024faiss} to execute high-dimensional vector similarity searches with low latency. The BAAI/bge-m3\cite{chen2024bge} model serves as the primary embedding backbone, encoding semantic states into dense vector representations.

\section{Prompt Templates}
\label{app:prompts}

This section presents the exact system prompts employed across the SGA-MCTS pipeline. These prompts act as the interface between our structured algorithms and the LLM's reasoning capabilities. In the templates below, `{{variable}}` denotes dynamic slots populated at runtime based on the execution context.

\subsection{SGA Extraction (De-lexicalization)}
An extractor model utilizes the following prompt to analyze raw MCTS trajectories. Its primary function is to abstract concrete execution paths into generalized, de-lexicalized State-Goal-Action (SGA) triplets, as described in Section \ref{sec:sga_extraction}.

\begin{table*}[ht]
\centering
\begin{tcolorbox}[title={System Prompt: SGA Extractor}]
\# ROLE \\
You are the \textbf{SGA Extractor} for an advanced autonomous agent. Your task is to analyze raw execution trajectories and distill them into generalized, atomic \textbf{State-Goal-Action (SGA)} patterns.
\\[1.5ex]
\# CONTEXT \\
The input contains a user request and a step-by-step trace of an agent using tools to solve it. Your goal is to extract \textbf{reusable logic} from this trace. These SGAs will be stored in a knowledge base to guide future agents. Therefore, the output must be \textbf{de-lexicalized} (stripped of specific values) and \textbf{templated}
\\[1.5ex]
\# 2. SCHEMA DEFINITIONS
\\[1.5ex]
\textbf{State (S)} \\
Describes the agent's understanding of the situation before acting.
\begin{itemize}
    \item \textbf{state\_summary}: A generic, high-level summary of the agent's current situation. It should describe the problem to be solved, focusing on what information is needed and what information is already available, without mentioning specific values.
\end{itemize}
\textbf{Goal (G)} \\
Describes the agent's immediate intent for the current step.
\begin{itemize}
    \item \textbf{goal}: The specific, high-level sub-goal the agent is trying to achieve. This should describe the question to be answered or the objective to be met at this stage, in a tool-agnostic way.
    \item Good: Find available flights matching the specified criteria.
    \item Good: Verify the current status of an order.
\end{itemize}
\textbf{Action (A)} \\
Describes the strategic category of the action taken to achieve the Goal.
\begin{itemize}
    \item \textbf{action}: A generalized description of the type of action the agent performs. This should be abstract and never mention the specific tool name. It describes the how in a strategic sense.
\end{itemize}
\# INPUT FORMAT \\
A JSON object containing the \texttt{question} and the \texttt{trajectory} (a list of actions and results).
\\[1.5ex]
\# OUTPUT FORMAT \\
You must output a strictly valid JSON object adhering to the following structure:
\\[1.5ex]
\# RULES \& CONSTRAINTS
\begin{enumerate}
    \item \textbf{Atomicity}: If a trajectory has 3 steps (A $\rightarrow$ B $\rightarrow$ C), output \textbf{3 separate SGA triplets}, not one combined chain. Each step is an independent training example.
\end{enumerate}
\end{tcolorbox}
\caption{The formal system prompt for the SGA Extraction. The rules ensure consistent extraction of reusable SGA patterns from execution traces.}
\label{tab:prompt_sga_extractor}
\end{table*}

\subsection{Reflection Generator}
The Reflection Generator prompt serves as a critic within the MCTS loop. It evaluates whether the current trajectory is logically sound and grounded in tool outputs, ensuring high-quality data for the experience store.

\begin{table*}[ht]
\centering
\begin{tcolorbox}[title={System Prompt: MCTS Messages Evaluation Expert}]
\# ROLE \\
You are a \textbf{Messages Evaluation Expert} specializing in analyzing Tool Learning / Agentic workflows. Your objective is to audit the logical connection between tool outputs and the AI's final answer.
\\[1.5ex]
\# CONTEXT \\
You will be presented with a conversation trace involving an AI and various Tools. The trace may include: \\
1. \textbf{AI Messages:} Tool calls or final answers. \\
2. \textbf{Tool Messages:} The raw results returned from a tool.
\\[1.5ex]
\# EVALUATION CRITERIA \\
You must verify if the \textbf{Final Answer} is logically derived from the \textbf{Tool Execution Results}.

\textbf{1. Completion Status} \\
* If the final message contains an \texttt{<answer>...</answer>} block and \textit{no} new tool calls, consider the task "Solved".

\textbf{2. Grounding Verification (The Core Task)} \\
Once the task is deemed "Solved," you must judge the \textbf{validity} based \textit{strictly} on the provided traces.
\begin{itemize}
    \item \textbf{High Score Criteria (Grounded):} The final answer is directly derived from the information provided in the \texttt{Tool: result} messages. The logic is traceable.
    \item \textbf{Low Score Criteria (Hallucinated/Ungrounded):} The final answer ignores or contradicts tool outputs, or appears to be generated solely from pre-trained knowledge.
\end{itemize}

\# SCORING RUBRIC \\
* \textbf{Pass / High Score:} The AI successfully used the tool data to construct the answer. \\
* \textbf{Fail / Low Score:} The AI generated an answer "by itself" without relying on the tool trace evidence.
\\[1.5ex]
\# EXAMPLE SCENARIOS \\
\textbf{Scenario A (High Score)}: Tool returns \texttt{\{"temp": "15C"\}}. AI answers \texttt{<answer> 15C </answer>}. Verdict: Fully supported. \\
\textbf{Scenario B (Low Score)}: Tool returns \texttt{\{"temp": "15C"\}}. AI answers \texttt{<answer> 30 </answer>}. Verdict: Not directly from tool result. \\
\textbf{Scenario C (Low Score)}: Tool returns \texttt{Error}. AI answers "The user is John Doe". Verdict: Severe hallucination.
\\[1.5ex]
\# TASK \\
Analyze the provided trace and provide your evaluation.
\\[1.5ex]
\{\{format\_prompt\}\}
\end{tcolorbox}
\caption{The formal system prompt for the Reflection Generator. This module acts as a critic to ensure that the agent's final output is explicitly grounded in the observation history rather than parametric memory.}
\label{tab:prompt_trace_eval}
\end{table*}

\subsection{Online Decision Maker}
During the online phase, the Decision Maker uses the following prompt to arbitrate between retrieved experiences and the current context, functioning as the primary actuator of the system.

\begin{table*}[ht]
\centering
\begin{tcolorbox}[title={System Prompt: Decision Maker}]
\# ROLE \\
You are a professional agent in an autonomous agent system. Your role is to decide which tool to call and what parameters to use.
\\[1.5ex]
\# CONTEXTUAL INPUTS \\
- Question: \{\{question\}\} \\
- Retrieved Experiences: \{\{experiences\_text\}\}
\\[1.5ex]
\# OPERATIONAL CONSTRAINTS \\
* ACTION\_MANDATORY: You MUST invoke at least one tool. Do not just respond with text unless you can answer the question based on the tool response. \\
* NO\_HISTORY\_DUPLICATES: You MUST NOT repeat a tool call with the exact same parameters used previously in this conversation history. \\
* NO\_USER\_CLARIFICATION: Do not ask questions back to the user. You must infer needed information and proceed with an attempt. \\
* ERROR\_RECOVERY: If a previous tool call failed, DO NOT retry it immediately. Change arguments significantly based on the feedback. \\
* STEP\_BY\_STEP: ALWAYS attempt to decompose the task and solve it sequentially using the available tools.
\end{tcolorbox}
\caption{The formal system prompt for the Decision Maker. The constraints are designed to minimize hallucination and ensure logical tool-chaining in zero-shot scenarios.}
\label{tab:prompt_decision_agent}
\end{table*}

\subsection{SGA Retriever Planner}
The SGA Retriever Planner analyzes the current execution state to formulate precise queries for the experience store.

\begin{table*}[ht]
\centering
\begin{tcolorbox}[title={System Prompt: SGA Retriever Planner}]
\# ROLE \\
You are the \textbf{SGA Retriever Planner} of an autonomous agent system. Your role is to analyze the current situation, extract state information, and prepare queries for SGA experience retrieval.
\\[1.5ex]
\# CONTEXTUAL INPUTS \\
- User Request: \{\{question\}\} \\
- Execution History: \{\{history\_str\}\} \\
- Current World Model (Known Info): \{\{current\_known\}\}
\\[1.5ex]
\# TASK \\
1. \textbf{Update Known Info}: Extract new facts from execution history and merge with existing knowledge \\
2. \textbf{State Analysis}: Generate abstract state summary suitable for semantic retrieval of similar past experiences \\
3. \textbf{Goal Definition}: Identify the immediate next goal to achieve \\
4. \textbf{Slot Extraction}: List available symbolic slots (e.g., <CITY>, <DATE>, <ID>)
\\[1.5ex]
\# OUTPUT REQUIREMENTS \\
Provide structured output with these fields:
\begin{itemize}
    \item \textbf{thought}: Your reasoning process and current situation analysis
    \item \textbf{updated\_known\_info}: Dictionary of new facts from history (can be empty)
    \item \textbf{state\_summary}: Abstract state description for SGA retrieval (avoid specific values, focus on patterns)
    \item \textbf{available\_slots}: List of symbolic slot tags available in current context
    \item \textbf{next\_goal}: Immediate actionable goal for the next step
\end{itemize}
\# EXAMPLES \\
For "What's weather in Beijing?":
\begin{itemize}
    \item state\_summary: "User requests weather information for a specific location"
    \item available\_slots: ["<LOCATION>"] 
    \item next\_goal: "Get weather data for specified location"
\end{itemize}
\# LANGUAGE \\
English by default
\# NOTE \\
You are NOT responsible for task completion decisions - focus on state analysis and goal formulation.
\end{tcolorbox}
\caption{The formal system prompt for the SGA Retriever Planner. The structured format ensures consistent state analysis and retrieval query preparation.}
\label{tab:prompt_sga_planner}
\end{table*}

\begin{figure}[t]
    \centering
    \includegraphics[width=0.95\linewidth]{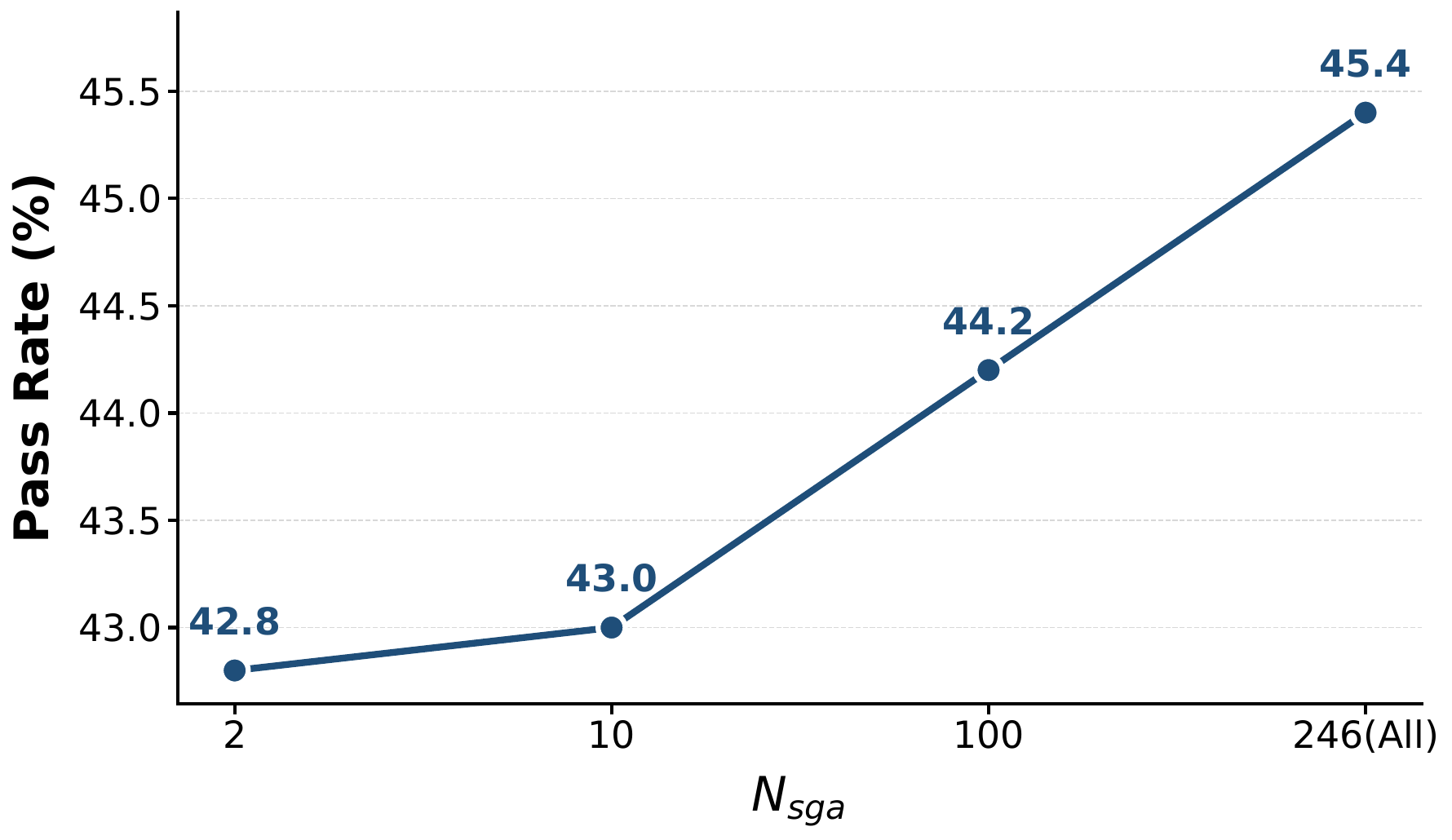}
    \caption{Impact of Experience Volume. Performance improves from 42.3\% ($N=2$) to 45.4\% ($N=246$). The high starting point highlights the data efficiency of SGA atoms, while the sustained growth demonstrates the value of broader coverage.}
    \label{fig:sga_scaling}
\end{figure}

\section{Additional Experimental Results}
\label{app:additional_results}

\subsection{Data Efficiency and Store Scaling}
We further analyze the sensitivity of the agent's performance to the scale of the atomic experience store ($N_{SGA}$). Figure \ref{fig:sga_scaling} plots the success rate as a function of the number of stored atoms.

The results exhibit a logarithmic growth trajectory. Performance climbs rapidly from a robust baseline of 42.3\% at $N=2$ to a peak of 45.4\% at $N=246$. This saturation profile highlights the high information density of our SGA abstraction: a relatively small core of canonical atoms is sufficient to capture the universal reasoning logic of the domain. The subsequent marginal gains suggest that expanding the repository primarily helps in resolving long-tail edge cases rather than learning fundamental capabilities.

\subsection{Qualitative Analysis of the Inference Workflow.}
Table \ref{tab:sga_inference_process} illustrates the complete lifecycle of an atomic reasoning step.

As shown in the top block, the Stored SGA Atom serves as a generic logic template. It is de-lexicalized, containing only the schema of required information (e.g., required\_slots: ["<YEAR>", "<GENRE>"]) and an abstract action definition, decoupled from specific entity values.

The bottom block details the Online Inference, which proceeds in a "Plan-then-Ground" manner: \begin{itemize} \item \textbf{Planning Phase}: Upon receiving the user query ("cartoons from '94"), the agent first analyzes the context to generate a semantic sub\_goal and extracts concrete values into candidate\_slots (mapping "cartoons" to animated and "'94" to 1994). \item \textbf{Retrieval and Grounding}: The system uses the generated goal and slots to query the experience store. Upon retrieving the matching SGA (th\_filter\_constraint\_01), the agent instantiates the abstract action template with the concrete values, resulting in the final executable tool call. \end{itemize} This mechanism ensures that the agent follows proven reasoning patterns (from MCTS) while dynamically adapting to new data instances.

\subsection{Tool Schema Specification}
To understand the complexity of the environment, we present the JSON definitions of the meta-cognitive operators injected into the agent's action space. Table \ref{tab:meta_operators_refined} details the schemas for \texttt{Plan} and \texttt{Reflection}. These operators are not external APIs but internal cognitive scaffolds designed to guide the MCTS exploration process.

\subsection{Meta-Cognitive Operators in MCTS}
\label{app:mcts_algo}
To address the reviewer's inquiry regarding how \texttt{Plan} and \texttt{Reflect} guide the tree construction, Algorithm \ref{alg:mcts_meta} illustrates their integration during the node expansion phase. These meta-actions are modeled as internal LLM calls that prune the vast tool space before actual execution.

\begin{algorithm}[h]
\caption{Meta-Cognitive Node Expansion in MCTS}
\label{alg:mcts_meta}
\begin{algorithmic}[1]
\REQUIRE Current state $s_t$, Global Goal $g$, Action Space $\mathcal{A} = \mathcal{A}_{tool} \cup \{\text{Plan}, \text{Reflect}\}$
\IF{is\_initial\_state($s_t$) \textbf{or} sub-goal\_completed($s_t$)}
    \STATE \textcolor{gray}{// Invoke Plan operator to decompose tasks}
    \STATE $g_{next} \leftarrow \text{LLM\_Call}(\texttt{Plan}, s_t, g)$
    \STATE Prune $\mathcal{A}_{tool}$ to retain only tools relevant to $g_{next}$
\ENDIF
\STATE Sample and execute tool action $a_t \sim \pi(a \mid s_t, g_{next})$
\STATE Obtain observation $o_t$ and calculate reward $r_t$ (Eq. 1)
\IF{$r_t == 0$ (Execution Failure)}
    \STATE \textcolor{gray}{// Invoke Reflect operator for error recovery}
    \STATE $pivot\_strategy \leftarrow \text{LLM\_Call}(\texttt{Reflect}, s_t, o_t)$
    \STATE Backpropagate penalty and update UCB values
    \STATE Prioritize $pivot\_strategy$ in the next iteration
\ELSE
    \STATE Append $(s_t, g_{next}, a_t)$ to valid trajectory $\tau$
\ENDIF
\end{algorithmic}
\end{algorithm}

\begin{table*}[ht]
\centering

\begin{tcolorbox}[
    title={Stored SGA Atom (De-lexicalized Logic)},
    fonttitle=\bfseries\small,
    colback=gray!5,
    colframe=green!40!black
]
\small\ttfamily
\{ \\
\hspace*{1em}"sga\_id": "th\_filter\_constraint\_01", \\
\hspace*{1em}"entry": \{ \\
\hspace*{2em}"state": \{ \\
\hspace*{3em}"description": "Requires filtering entity candidates via temporal and genre constraints.", \\
\hspace*{3em}"required\_slots": [ "<YEAR>", "<GENRE>" ] \\
\hspace*{2em}\}, \\
\hspace*{2em}"goal": "Retrieve candidate list matching constraints.", \\
\hspace*{2em}"action": "wiki\_search" \\
\hspace*{1em}\} \\
\}
\end{tcolorbox}

\vspace{0.4cm}

\begin{tcolorbox}[
    title={Online Inference Workflow (Planning $\rightarrow$ Retrieval $\rightarrow$ Grounding)},
    fonttitle=\bfseries\small,
    colback=gray!5,
    colframe=teal!60!black
]
\small\ttfamily
\{ \\
\hspace*{1em}"user\_query": "Find me some cartoons from '94.", \\
\hspace*{1em}"\textbf{phase\_1\_planning}": \{ \\
\hspace*{2em}"thought": "I need to find films. The user specifies a year and a genre.", \\
\hspace*{2em}"generated\_sub\_goal": "Retrieve candidate list matching constraints.", \\
\hspace*{2em}"extracted\_slots": \{ \\
\hspace*{3em}"<YEAR>": "1994", \\
\hspace*{3em}"<GENRE>": "animated" \\
\hspace*{2em}\} \\
\hspace*{1em}\}, \\
\hspace*{1em}"\textbf{phase\_2\_grounding}": \{ \\
\hspace*{2em}"retrieved\_sga": "th\_filter\_constraint\_01", \\
\hspace*{2em}"reasoning": "The retrieved SGA aligns with my sub-goal. I will fill its abstract action template with my extracted slots.", \\
\hspace*{2em}"final\_tool\_call": \{ \\
\hspace*{3em}"tool": "wiki\_search", \\
\hspace*{3em}"arguments": (year=1994, genre=animated) \\
\hspace*{2em}\} \\
\hspace*{1em}\} \\
\}
\end{tcolorbox}

\caption{\textbf{Qualitative Example of the Retrieval-Augmented Planning Workflow.} 
The process operates in two stages: 
(1) The \textbf{Stored SGA Atom} (top) represents a frozen, de-lexicalized reasoning primitive residing in the experience store $\mathcal{D}$. Its \texttt{required\_slots} define the schema needed for activation.
(2) During the \textbf{Online Inference Workflow} (bottom), the agent first performs \textit{Planning} to generate a sub-goal and extract candidate slot values (e.g., mapping "'94" to \texttt{<YEAR>}). This structured intent triggers the retrieval of the matching SGA. Finally, in the \textit{Grounding} phase, the agent instantiates the abstract logic with concrete values to execute the precise tool call.}
\label{tab:sga_inference_process}
\end{table*}

\begin{table*}[ht]
\centering

\begin{tcolorbox}[
    title={JSON Schema: Plan Operator },
    fonttitle=\bfseries\small,
    colback=gray!5,
    colframe=blue!75!black
]
\small\ttfamily
\{ \\
\hspace*{1em}"name": "plan", \\
\hspace*{1em}"description": "Decomposes complex objectives into actionable steps. Acts as 'Memory' to anchor the search branch.", \\
\hspace*{1em}"parameters": \{ \\
\hspace*{2em}"type": "object", \\
\hspace*{2em}"properties": \{ \\
\hspace*{3em}"task\_plan": \{ \\
\hspace*{4em}"type": "string", \\
\hspace*{4em}"description": "Structured breakdown: 1. Analysis; 2. Strategy; 3. Execution steps." \\
\hspace*{3em}\}, \\
\hspace*{3em}"priority\_focus": \{ \\
\hspace*{4em}"type": "string", \\
\hspace*{4em}"description": "The single most critical aspect to prioritize in the immediate next step." \\
\hspace*{3em}\} \\
\hspace*{2em}\}, \\
\hspace*{2em}"required": ["task\_plan"] \\
\hspace*{1em}\} \\
\}
\end{tcolorbox}

\vspace{0.4cm} 

\begin{tcolorbox}[
    title={JSON Schema: Reflection Operator (\texttt{reflection})},
    fonttitle=\bfseries\small,
    colback=gray!5,
    colframe=red!75!black
]
\small\ttfamily
\{ \\
\hspace*{1em}"name": "reflection", \\
\hspace*{1em}"description": "Critically evaluates the execution trace to identify logical flaws and brainstorm pivots.", \\
\hspace*{1em}"parameters": \{ \\
\hspace*{2em}"type": "object", \\
\hspace*{2em}"properties": \{ \\
\hspace*{3em}"current\_context": \{ "type": "string", "description": "Summary of observed state." \}, \\
\hspace*{3em}"critique": \{ "type": "string", "description": "Identification of potential flaws or edge cases." \}, \\
\hspace*{3em}"alternative\_ideas": \{ "type": "string", "description": "Proposed recovery paths if current branch fails." \} \\
\hspace*{2em}\}, \\
\hspace*{2em}"required": ["current\_context", "critique", "alternative\_ideas"] \\
\hspace*{1em}\} \\
\}
\end{tcolorbox}

\caption{\textbf{Meta-cognitive operator specifications derived from implementation.} The \texttt{task\_decomposition} (top) enforces a three-stage structural prior (Analysis, Strategy, Execution), while \texttt{reflection} (bottom) implements a mandatory multi-perspective critique to bypass local optima during MCTS exploration.}
\label{tab:meta_operators_refined}
\end{table*}

\end{document}